\documentclass[10pt,twocolumn,letterpaper]{article}

\usepackage{cvpr}              
\definecolor{cvprblue}{rgb}{0.21,0.49,0.74}
\usepackage[pagebackref,breaklinks,colorlinks,allcolors=cvprblue]{hyperref}

\def\paperID{XXXX} 
\def\confName{CVPR}
\def\confYear{2026}

\title{Gaze into the Details: Locality-Sensitive Enhancement for OCTA Retinal Vessel Segmentation}

\author{
	Tuopusen Huang, Ding Ma, Xiangqian Wu \thanks{corresponding author.}\\
	Faculty of Computing\\
	Harbin Institute of Technology\\
	Harbin, China\\
	{\tt\small tuopusen\_huang@stu.hit.edu.cn, madingcs@hit.edu.cn, xqwu@hit.edu.cn}
}

\begin{document}
\maketitle
\begin{abstract}
Existing deep learning frameworks for Optical Coherence Tomography Angiography (OCTA) vessel segmentation are largely derived from the U-Net architecture, which serves as the foundation for most current designs. However, most of these methods focus only on holistic representation, struggling to address the problem of low local contrast unique to OCTA, which leads to vessel discontinuities and loss of detail. To address these problems, we propose LSENet, which builds upon the U-Net architecture by introducing three core innovative modules: To address vessel discontinuities, we introduce the Patch Information Enhance module (PIE), which replaces standard skip connections to execute patch-wise attention. To mitigate detail loss, the Multiscale Feature Fusion module (MFF) is proposed to feed the PIE module rich, multi-scale information by extracting visually interpretable features from both the original input and preceding layers. Finally, the Connectivity Refinement Decoder (CRD) is designed to refine features from all levels and utilize a large kernel in the final convolutional layer to reduce fragmentation. Experiments on three public datasets (OCTA-500, ROSE-1, and ROSSA) demonstrate that our proposed LSENet achieves state-of-the-art performance while requiring fewer parameters.
\end{abstract}    
\section{Introduction}
\label{sec:intro}

Optical Coherence Tomography Angiography (OCTA) is a non-invasive imaging technique for high-resolution visualization of the ocular microvasculature \cite{1,2,3,4}, which holds significant value in the diagnosis and monitoring of pathologies such as diabetic retinopathy, macular degeneration, and glaucoma \cite{5,6,7}. Artificial Intelligence and Deep Learning are increasingly utilized to enable the automated and reproducible analysis of OCTA data, thereby improving screening, diagnosis, and treatment monitoring \cite{8}. However, achieving accurate retinal vessel segmentation faces multiple challenges \cite{9}: The low image Signal-to-Noise Ratio, coupled with the complex structure, diverse orientations, varied sizes of the retinal vasculature, and low local contrast make segmentation extremely difficult. Furthermore, the low local contrast leads to ambiguity in uncertain regions, such as microvessels, vessel edges, and endpoints, resulting in vessel discontinuities and loss of detail.

\begin{figure}[t]
	\setlength{\abovecaptionskip}{0.cm}
	\setlength{\belowcaptionskip}{-0.5cm}
	\centering
	\includegraphics[width=1.0\linewidth]{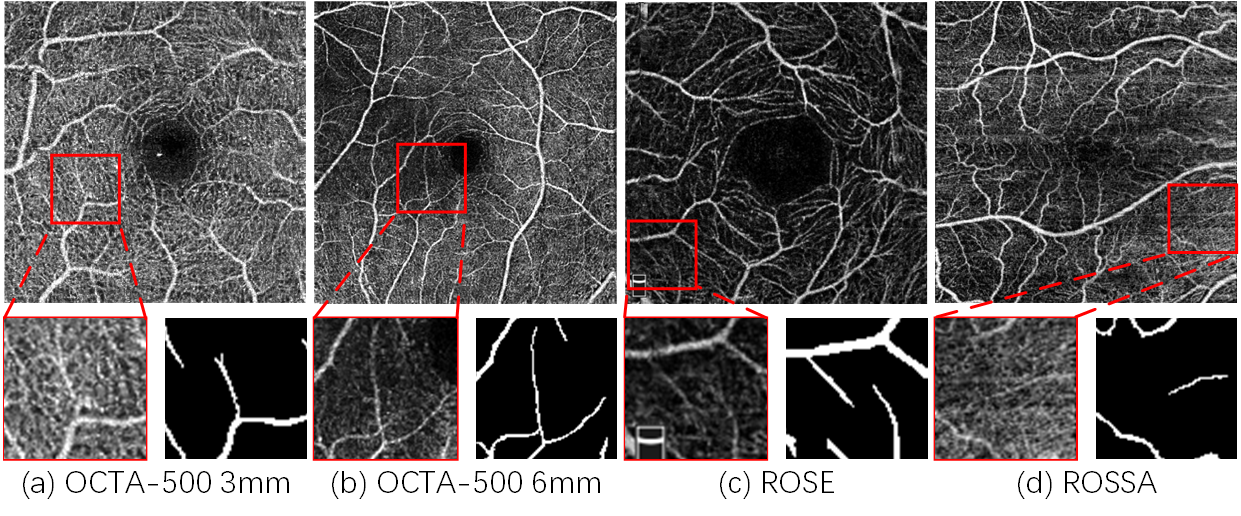}
	\caption{Examples of low local contrast challenges from the OCTA-500, ROSE-1, and ROSSA datasets. (a) illustrates low contrast in a bright region, while (b)-(d) show low contrast in dark regions. In all subfigures, the magnified local region (bottom left) shows vessels that are difficult to distinguish from the background, as confirmed by the ground truth (bottom right), posing a segmentation challenge.}
	\label{fig:Figure 1}
\end{figure}

In recent years, numerous deep learning studies have been proposed for retinal vessel segmentation in OCTA images to address the field's inherent challenges. \textbf{Customized U-Net architectures} \cite{14, 32} refine the standard U-Net structure to enhance vessel recognition specifically for the OCTA modality. \textbf{Transformer-based architectures} \cite{10, 11, 13} employ token-based processing, tokenizing vessel semantics to effectively capture global contextual information. Additionally, \textbf{specialized strategies for data optimization and label scarcity} \cite{15, 12} leverage specific training protocols or network designs to enhance the model's information capture capabilities.

Concurrently, most methods rely on holistic representation, effectively taking a ``glance'' at global features. While this approach excels at capturing the geometry of main vessels and preserving global spatial context, it struggles with the prevalent challenge of low local contrast in OCTA data (as shown in Fig.~\ref{fig:Figure 1}), especially at the distal ends of target vessels. Consequently, models relying on these holistic features often fail to segment the fine vascular endings accurately, leading to vessel discontinuities and a loss of detail. To address these limitations, we posit that processing these local regions individually, free from the interference of the global foreground and background, enables a more robust identification of these low contrast regions. Accordingly, we propose LSENet, a lightweight and efficient network that builds upon the holistic ``glance'' of its U-Net architecture by augmenting it with a locality-sensitive, patch-based module, enabling it to ``gaze into details'' by amplifying multi-scale feature information within localized regions.

Specifically, we strengthen the global features derived from the primary U-Net by incorporating local perception capabilities through three core modules. We propose a Patch Information Enhance module (PIE) that, instead of operating directly on global features, partitions the feature map into several patches. This allows the model to focus on extracting regions containing vessel information individually within each local patch. Furthermore, we introduce a Multiscale Feature Fusion module (MFF) to decompose the global features at multiple scales before they are fed into the PIE module. The MFF is designed to extract vessel features at various scales, subsequently fusing and selecting the most salient feature information, thereby supplying PIE with enhanced potential vascular details. While this additional vascular information helps the PIE module identify hard-to-detect, low-contrast target vessels, it may also introduce more false positives (refer to Sec.~\ref{subsec:Ablation Study}). To mitigate this risk, we propose a Connectivity Refinement Decoder (CRD) that utilizes a large kernel convolution as the final output layer to ensure the model achieves higher vessel connectivity.

The contributions of our study are summarized as follows:

\begin{itemize}
    \item We propose a novel Locality-Sensitive Enhancement Network (LSENet), which employs a patch-based processing methodology to ``gaze into details'' in OCTA vessel segmentation. To our knowledge, this network is the first in the OCTA vessel segmentation domain to leverage a locality-sensitive information enhancement strategy.
    \item We introduce three core modules that enable the model to emphasize vascular details: The MFF module extracts and fuses multi-scale features to overcome detail loss; the PIE module assigns patch-wise attention to handle vessel discontinuities; and the CRD uses a large convolution kernel on the refined information to ensure superior connectivity.
    \item We demonstrate the state-of-the-art performance of our method on the OCTA vessel segmentation task through extensive experiments on three publicly available datasets (ROSE-1 \cite{42}, OCTA-500 \cite{41}, and ROSSA \cite{12}).
\end{itemize}

\section{Related Works}
\label{sec:Related Works}

\begin{figure*}
	\setlength{\abovecaptionskip}{0.cm}
	\setlength{\belowcaptionskip}{-0.5cm}
	\centering
	\includegraphics[width=1\linewidth]{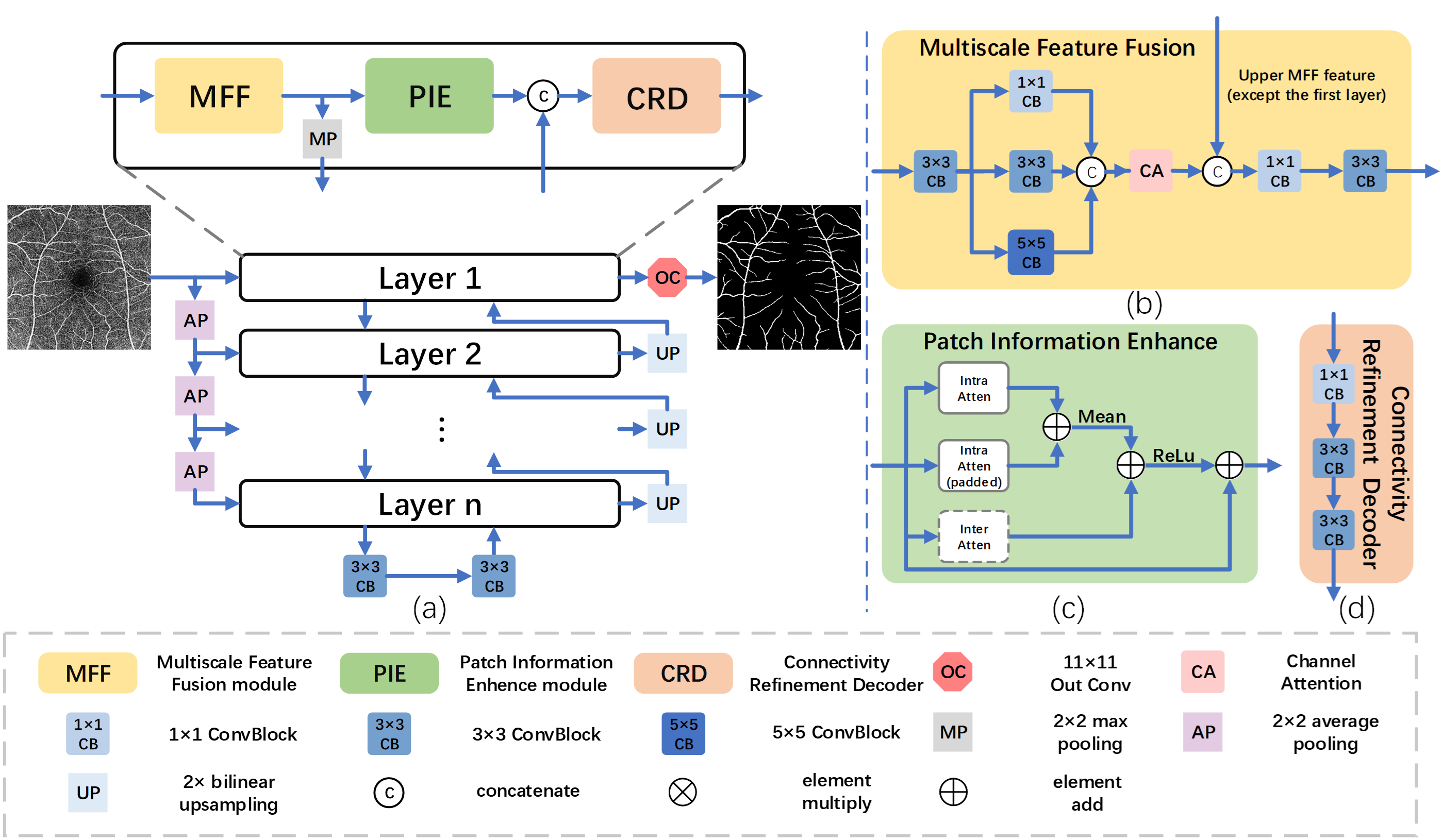}
	\caption{Architecture of LSENet and its core modules. (a) The main architecture, composed of stacked layers (each with three modules). (b) Multiscale Feature Fusion (MFF) module: Fuses vessel features using multiscale kernels and channel attention. (c) Patch Information Enhance (PIE) module: Handles vessel discontinuities via intra- and inter-patch attention. (d) Connectivity Refinement Decoder (CRD): Refines layer-wise features and, aided by a final large-kernel convolution, maintains vessel connectivity. }
	\label{fig:Figure 2}
\end{figure*}

\subsection{Retina Vessel Segmentation}

Retina vessel segmentation, historically focused on Color Fundus Photography (CFP), was long dominated by the U-Net architecture \cite{18}. This framework was progressively refined with enhanced context, attention, and boundary modeling \cite{16, 19, 20, 24}; adaptations for complex morphology via deformable convolutions and multi-scale interaction \cite{21, 22, 23, 26, 27}; and more efficient architecture \cite{25}.

These foundational studies informed subsequent work on OCTA, which has emerged as a mainstream retinal vessel examination method owing to its high-resolution, 3D microvascular details \cite{1,2,3,4}. However, these advanced characteristics also introduce novel challenges to vessel segmentation, such as intricate structures and low local contrast, rendering methodologies from previous modalities suboptimal. Consequently, OCTA-specific methods have been developed, including: \textbf{customized U-Net architectures} that employ dual-branch designs, directional convolutions, or multi-task learning \cite{14, 32, 38, 40}; \textbf{Transformer architectures} \cite{28} that leverage hybrid CNN-Transformer models or adapted foundation models \cite{10, 11, 13}; and \textbf{strategies targeting data optimization} through multi-modal fusion, 3D-to-2D projection, and semi-supervised or unsupervised approaches \cite{12, 15, 35, 36, 37, 39}.

\subsection{Local Enhancement Methods}

Strategies for local enhancement, wherein an image is partitioned into patches or regions to extract fine-grained details, represent a pivotal research area in both natural and medical image processing.

\textbf{In the natural image domain}, this paradigm is applied across diverse tasks. For image classification, ViT \cite{43} partitions images into patch-based tokens. For semantic segmentation, the Swin Transformer \cite{30} computes self-attention within hierarchical shifted windows''. For object detection, the Twins model \cite{44} combines Locally-grouped (LSA) and Global Sub-sampled (GSA) attention. For facial recognition, PACVT \cite{47} employs a Patch Attention Unit'' (PAU), and for image captioning, the Patch Matters model \cite{45} uses a divide-then-aggregate'' strategy. Additionally, local feature en-hancement is critical for processing sparse spatio-temporal signals \cite{ren2024spikepoint,10946204}. \textbf{In the medical imaging domain}, this methodology is widely adopted for localized pathology analysis. For organ segmentation, Swin-Unet \cite{29} adapts the Swin Transformer's shifted window'' mechanism. For Alzheimer's disease diagnosis, DA-MIDL \cite{48} utilizes Patch-Nets'' with spatial attention. For breast tumor classification, SGLA-Net \cite{49} integrates global and local representations via a Global-Local Feature Interaction'' (GLFI) block. For melanoma analysis, GPLE \cite{46} introduces a ``Supplement Shifted Window Partition'' (SSWP).

Despite the widespread adoption of these local enhancement strategies in other domains, to the best of our knowledge, this paradigm has not yet been applied to the specific task of vessel segmentation within the OCTA domain. Our work, therefore, aims to address this research gap.
\section{Methodology}
\label{sec:Methodology}

\subsection{Overall Framework}
We propose the Locality-Sensitive Enhancement Network (LSENet) to address the challenges posed by low local contrast in OCTA images. As shown in Fig.~\ref{fig:Figure 2}, the LSENet architecture is built on a lightweight U-Net backbone, modified with a constant 64-channel count to reduce parameters.

Our core contributions are three modules that replace U-Net components: the \textbf{Multiscale Feature Fusion module (MFF)}, which extracts rich, multi-scale features from the raw input at various resolutions; the \textbf{Patch Information Enhance module (PIE)}, which replaces standard skip connections and employs patch-based processing to perform targeted local information enhancement in challenging regions; and the \textbf{Connectivity Refinement Decoder (CRD)}, a lightweight decoder that refines features from the PIE module and upsampling path, cooperating with a final large-kernel convolution to ensure vessel connectivity.

\begin{figure}
	\setlength{\abovecaptionskip}{0.cm}
	\setlength{\belowcaptionskip}{-0.5cm}
	\centering
	\includegraphics[width=0.8\linewidth]{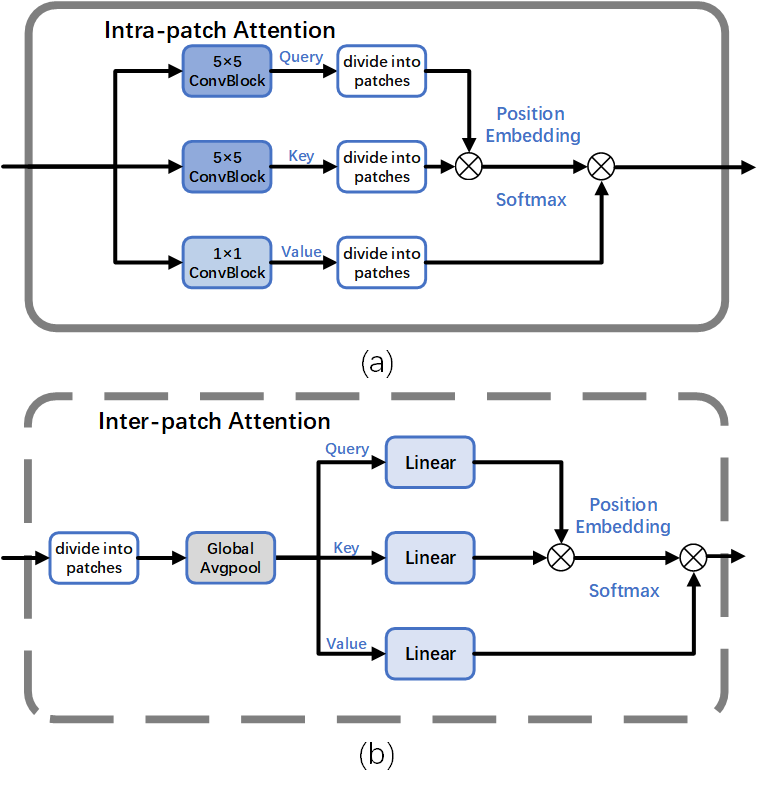}
	\caption{Components of the PIE module. (a) Intra-patch Attention: Enhances vessel features within patches. (b) Inter-patch Attention: Compensates for global information loss between patches.}
	\label{fig:Figure 3}
\end{figure}

\begin{figure}
  \centering
  \includegraphics[width=0.8\linewidth]{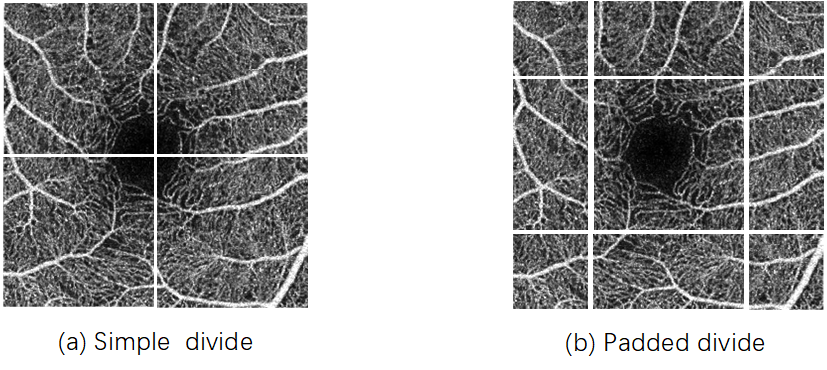}
  \caption{Interleaved partitioning for PIE intra-patch attention, illustrating (a) the simple (non-shifted) grid and (b) the shifted (padded) grid. Combining these overlapping strategies mitigates the limitations of a single partition, enhancing segmentation robustness. (Note: Patches enlarged for clarity).}
  \label{fig:Figure 4}
\end{figure}

\subsection{Patch Information Enhance Module (PIE)}

To explicitly capture local context while avoiding global extreme values, PIE (Fig.~\ref{fig:Figure 2}) replaces standard skip connections with a dual-branch patch-wise attention mechanism:

\textbf{Intra-Patch Attention (Fig.~\ref{fig:Figure 3} a):} Given an input $X \in \mathbb{R}^{C \times H \times W}$, we symmetrically pad it to be divisible by patch size $P$. Query ($Q_{intra}$), Key ($K_{intra}$), and Value ($V_{intra}$) are generated via convolutions ($5 \times 5$ for $Q_{intra}, K_{intra}$ to capture spatial priors; $1 \times 1$ for $V_{intra}$) and subsequently reshaped into non-overlapping patches of shape $\mathbb{R}^{N \times C \times P^2}$, where $N = \frac{H}{P} \times \frac{W}{P}$. By treating the $P^2$ pixels as the sequence length, attention is computed independently within each patch using a relative position encoding $P_{pos}$:
\begin{equation}
	\text{Attn}_{intra} = \text{Attn}(Q_{intra}, K_{intra}, V_{intra})
	\label{eq:attn_intra}
\end{equation}
\begin{equation}
	\text{Attn}(Q, K, V) = \text{Softmax}\left(\frac{QK^T + P_{pos}}{\sqrt{C}}\right)V
	\label{eq:attn}
\end{equation}

\begin{figure}[t]
	\setlength{\abovecaptionskip}{0.cm}
	\setlength{\belowcaptionskip}{-0.4cm}
	\centering
	\includegraphics[width=1\linewidth]{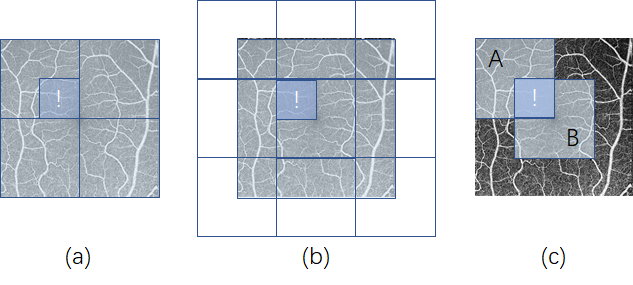}
	\caption{Receptive field comparison. (a) Original partitioning. (b) Shifted partitioning (padded by $P/2$). (c) The combined strategy doubles the effective receptive field.}
	\label{fig:Figure 5}
\end{figure}

\textbf{Shifted Intra-Patch Attention:} To prevent continuous vessels from being artificially severed by fixed boundaries, a parallel branch applies the identical $\text{Patch}(\cdot)$ and attention operations to $X$ after padding it by $P/2$ on all sides. The output is then cropped back to its original size ($\text{Attn}_{\text{intra, padding}}$). This shifted window approach enables cross-patch information flow, doubling the effective receptive field (Fig.~\ref{fig:Figure 5}).

\textbf{Inter-Patch Attention (Fig.~\ref{fig:Figure 3} b):} To recapture necessary global context, we apply spatial average pooling to each patch, yielding $N$ tokens. Linear layers generate $Q, K, V$, and standard attention is computed across the $N$ patches. The output ($\text{Attn}_{\text{inter}}$) is reshaped and bilinearly interpolated back to $H \times W$ .

The final feature $X_{PIE}$ fuses the three branches, followed by a ReLU activation and a residual connection:
\begin{equation}
	\text{Attn}_{\mathrm{X}}=\frac{\text{Attn}_{\text{intra}}+\text{Attn}_{\text{intra, padding}}}{2}+\text{Attn}_{\mathrm{inter}}
\end{equation}
\begin{equation}
	X_{PIE} = \text{ReLU}\left( \text{Attn}_{X}\right) + X
	\label{eq:pie_final}
\end{equation}

\begin{figure}[t]
	\setlength{\abovecaptionskip}{0.cm}
	\setlength{\belowcaptionskip}{-0.4cm}
	\centering
	\includegraphics[width=1\linewidth]{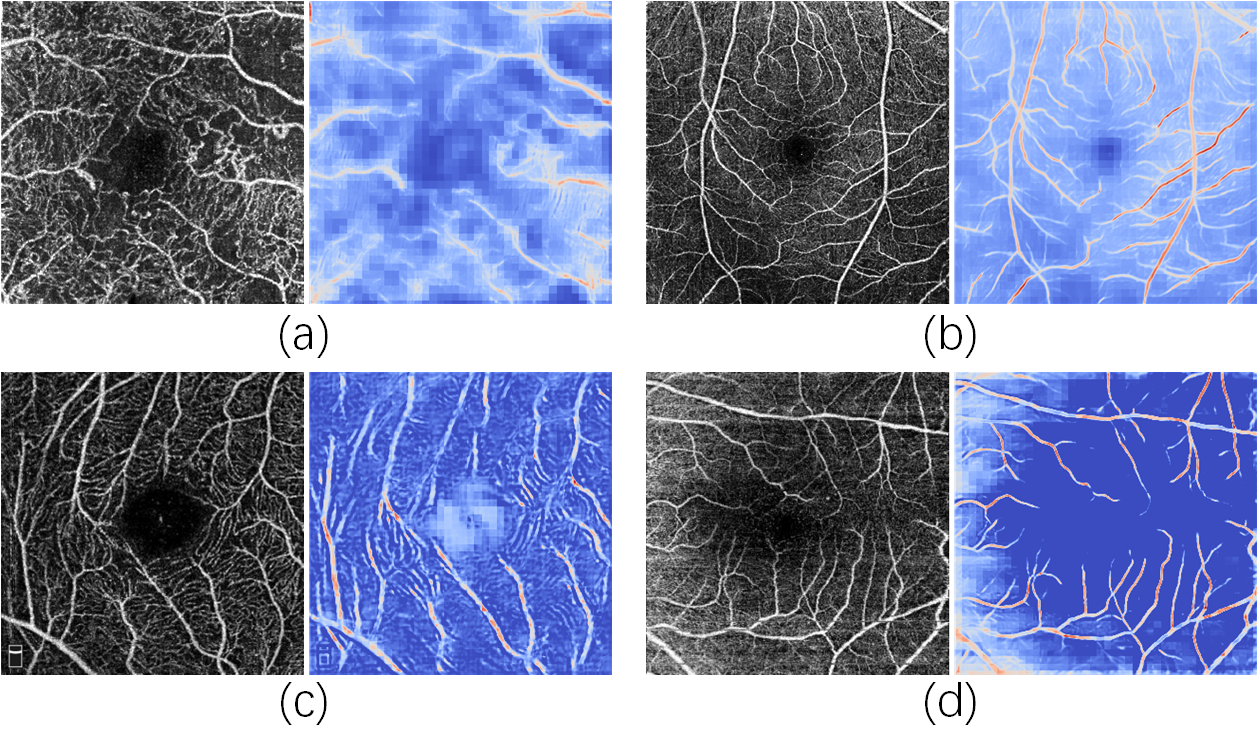}
	\caption{Attention heatmaps demonstrating PIE's focus. High attention (red) strongly aligns with target vessels compared to the background (blue).}
	\label{fig:Figure 6}
\end{figure}

As validated in Fig.~\ref{fig:Figure 6}, the module's attention heatmaps confirm that PIE functions as an effective local enhancer, providing refined, discriminative vessel features for the decoder.

\subsection{Multiscale Feature Fusion Module (MFF)}
The MFF module (Fig.~\ref{fig:Figure 2}) is designed to feed rich, interpretable features to the PIE module. It takes two inputs: $X_{down}$ (the output from the previous MFF layer) and $X_{in}$ (the raw input image downsampled via average pooling).

The $X_{in}$ input first passes through a $3 \times 3$ convolution block ($X_{pre}$, Eq.~\ref{eq:x_pre}), followed by parallel $1 \times 1$, $3 \times 3$, and $5 \times 5$ convolution  blocks to extract multi-scale features (Eqs.~\ref{eq:x_1}). In $X_{n \times n}$, $n \times n$ denotes the size of the convolution kernel. Specifically, the convolution  block (ConvBlock) includes one convolution, one GroupNorm, and one ReLU.
\begin{equation}
X_{pre} = \text{ConvBlock}_{3 \times 3}(X_{in})
\label{eq:x_pre}
\end{equation}
\begin{equation}
X_{n \times n} = \text{ConvBlock}_{n \times n}(X_{pre})
\label{eq:x_1}
\end{equation}

These features ($X_{multi}$) are screened by a Channel Attention (CA) \cite{19} module ($X_{att}$, Eqs.~\ref{eq:x_multi}-\ref{eq:x_att}). $X_{att}$ is then concatenated with $X_{down}$ (Eq.~\ref{eq:x_concat}) and fused via a $1 \times 1$ conv (Eq.~\ref{eq:x_fusion}) and a $3 \times 3$ conv (Eq.~\ref{eq:x_out}) to produce the final 64-channel output.
\begin{equation}
X_{multi} = \text{Concat}(X_{1 \times 1}, X_{3 \times 3}, X_{5 \times 5})
\label{eq:x_multi}
\end{equation}
\begin{equation}
X_{att} = \text{CA}(X_{multi})
\label{eq:x_att}
\end{equation}
\begin{equation}
X_{concat} = \text{Concat}(X_{att}, X_{down})
\label{eq:x_concat}
\end{equation}
\begin{equation}
X_{fusion} = \text{ConvBlock}_{1 \times 1}(X_{concat})
\label{eq:x_fusion}
\end{equation}
\begin{equation}
X_{MFF} = \text{ConvBlock}_{3 \times 3}(X_{fusion})
\label{eq:x_out}
\end{equation}

\begin{figure}[t]
  \setlength{\abovecaptionskip}{0.cm}
  \setlength{\belowcaptionskip}{-0.4cm}
  \centering
  \includegraphics[width=1\linewidth]{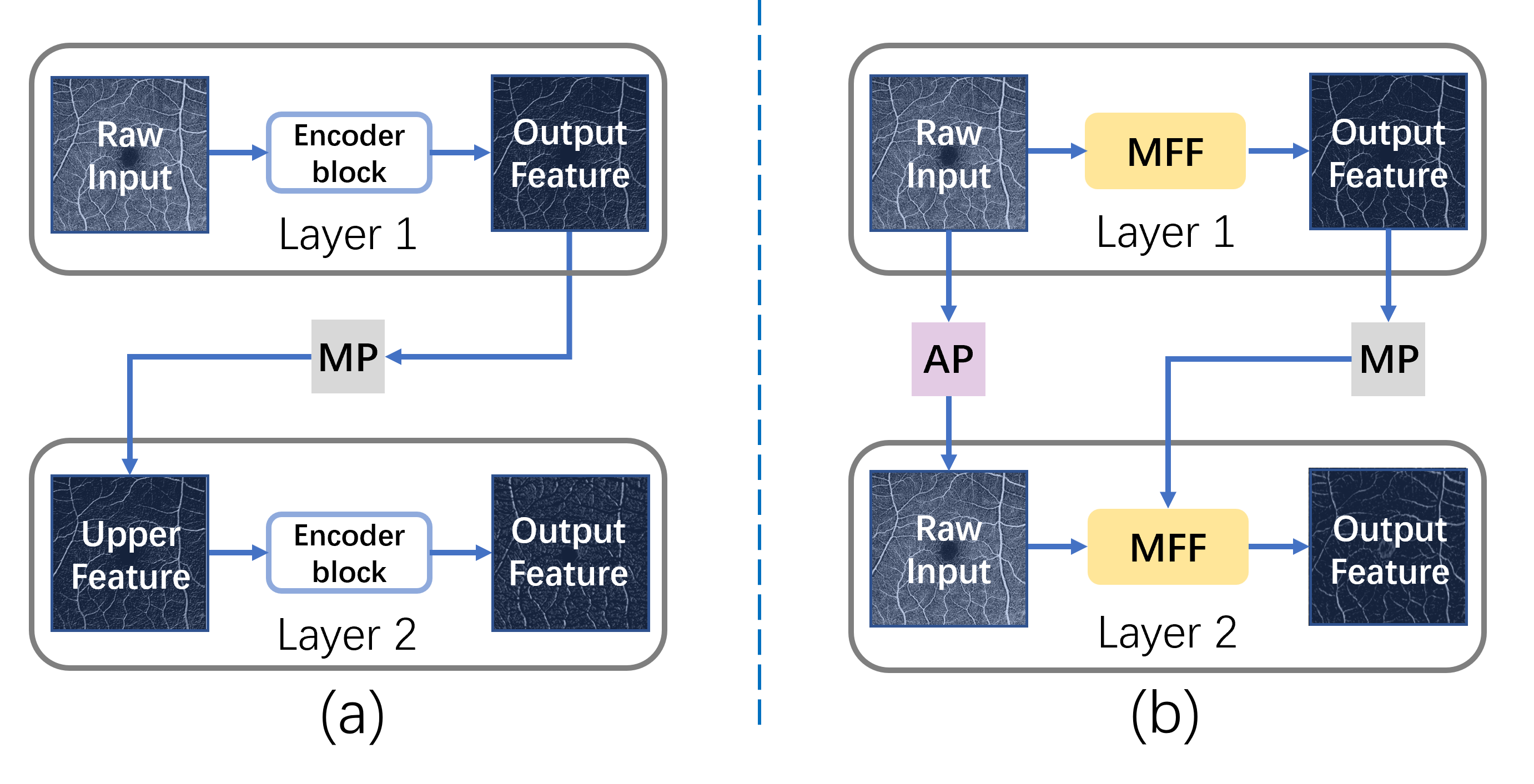}
  \caption{Input strategy comparison. (a) A standard U-Net block receives only max-pooled features from the preceding layer. (b) Our MFF module additionally receives the average-pooled raw input, a design that preserves visual interpretability at each stage.}
  \label{fig:Figure 7}
\end{figure}

Critically, to ensure our PIE module operates on meaningful local structures while maintaining visual interpretability, we use the downsampled raw image $X_{in}$ as the primary input, rather than only the abstract features from former encoder layers (Fig.~\ref{fig:Figure 7}).

Furthermore, to ensure robustness to the small batch sizes prevalent in medical imaging, all convolutional blocks in our network employ GroupNorm (8 channels/group) instead of BatchNorm.

\subsection{Connectivity Refinement Decoder (CRD)}

Our Connectivity Refinement Decoder (CRD) (Fig.~\ref{fig:Figure 2}) is a lightweight component designed to maintain a 64-channel output. At each stage, its core function is to refine upsampled features $X_{up}$ by fusing them with incoming PIE features $X_{PIE}$ using a $1 \times 1$ convolution block (Eq.~\ref{eq:x_refine}). This intermediate representation, $X_{refine}$, is then processed through two consecutive $3 \times 3$ convolution blocks to produce the stage's output, $X_{CRD}$ (Eq.~\ref{eq:x_CRDout}).
\begin{equation}
	X_{refine} = \text{ConvBlock}_{1 \times 1}(\text{Concat}(X_{up}, X_{PIE}))
	\label{eq:x_refine}
\end{equation}
\begin{equation}
	X_{CRD} = \text{ConvBlock}_{3 \times 3}(\text{ConvBlock}_{3 \times 3}(X_{refine}))
	\label{eq:x_CRDout}
\end{equation}

\begin{figure}
	\setlength{\abovecaptionskip}{0.cm}
	\setlength{\belowcaptionskip}{-0.5cm}
	\centering
	\includegraphics[width=0.8\linewidth]{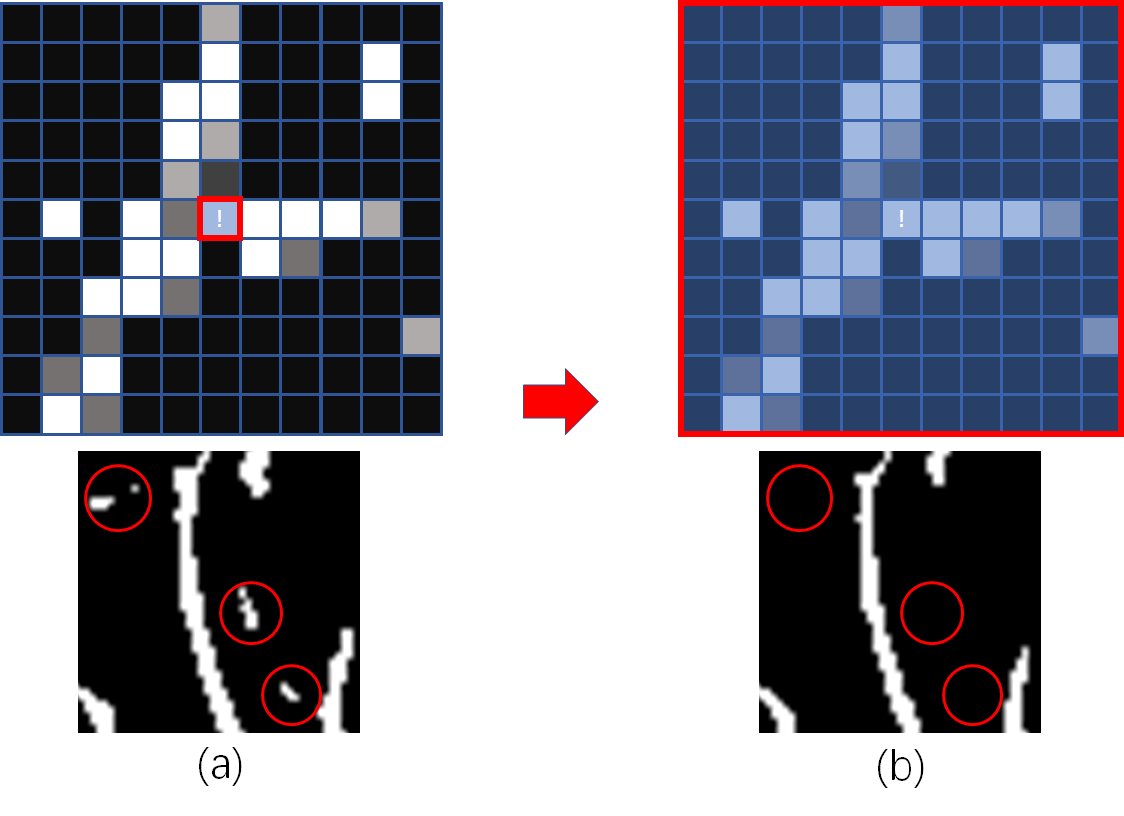}
	\caption{Improved Output Convolution (OutConv) comparison. (a) A standard U-Net's $1 \times 1$ kernel has a receptive field limited to a single point. (b) LSENet's $11 \times 11$ large kernel captures broader vascular context, significantly improving vessel connectivity and reducing isolated false positives.}
	\label{fig:Figure 8}
\end{figure}

The most critical component is the final output layer. We replace the standard $1 \times 1$ `OutConv` with a large $11 \times 11$ no bias convolution (Fig.~\ref{fig:Figure 8}). This is motivated by the continuous nature of vessels; segmenting a pixel requires contextual information from a larger neighborhood. While this introduces minor border artifacts, experiments (Sec.~\ref{subsec:Ablation Study}) confirm it significantly improves vessel connectivity and reduces fragmented endpoints.

\subsection{Loss function}
We employ a combined loss function, $Loss = Loss_{bce} + Loss_{dice}$, where the Binary Cross-Entropy (BCE) Loss (Eq.~\ref{eq:loss_bce}) guides the model in learning the overall vessel distribution (particularly during early training), while the DiceLoss (Eq.~\ref{eq:loss_dice}) focuses on optimizing the fine-grained segmentation of vessel details and endpoints.
\begin{equation}
	Loss_{bce} = - \frac{1}{n} \sum_{i=1}^{n} [y_i \log(\hat{y}_i) + (1 - y_i) \log(1 - \hat{y}_i)]
	\label{eq:loss_bce}
\end{equation}
\begin{equation}
Loss_{dice} = 1 - \frac{2 \sum_{i=1}^{n} y_i \hat{y}_i}{\sum_{i=1}^{n} y_i + \sum_{i=1}^{n} \hat{y}_i}
\label{eq:loss_dice}
\end{equation}
Here, $\hat{y}_i$ is the $i$-th pixel's predicted value, and $y_i$ is its ground truth label.

\begin{table*}
	\setlength{\abovecaptionskip}{0.cm}
	\setlength{\belowcaptionskip}{-0.5cm}
	\centering
	\scriptsize
	\caption{Quantitative comparison results for retinal vessel segmentation on the OCTA-500, ROSE-1 and ROSSA datasets (Mean ± SD). \textbf{Bold} indicates the best performance, \underline{underline} indicates the second best, and \underline{\underline{double underline}} indicates the third best. ``↑'' means the higher the better, while ``↓'' means the lower the better.}
	\label{tab:Table 1}
	\begin{tabular}{ccccccccc}
		\toprule
		Dataset & Model & Venue(year) & Dice\% ↑ & Sensitivity\% ↑ & Specificity\% ↑ & FDR\% ↓ & Accuracy\% ↑ & Kappa\% ↑ \\
		\midrule
		& UNet & MICCAI(2015) & 91.02 ± 0.06 & 90.30 ± 0.94 & \underline{\underline{99.41 ± 0.08}} & \underline{\underline{8.23 ± 0.92}} & 98.80 ± 0.01 & 90.38 ± 0.07 \\
		& CENet & TMI(2019) & 89.33 ± 0.05 & 89.31 ± 0.37 & 99.23 ± 0.03 & 10.65 ± 0.31 & 98.56 ± 0.01 & 88.55 ± 0.05 \\
		& VesselNet & MICCAI(2019) & 90.20 ± 0.06 & 89.01 ± 0.15 & 99.40 ± 0.01 & 8.58 ± 0.10 & 98.69 ± 0.01 &  89.50 ± 0.07 \\
		& CSNet & MedIA(2021) & 91.11 ± 0.05 & \underline{\underline{90.92 ± 0.56}} & 99.37 ± 0.05 & 8.69 ± 0.58 & 98.80 ± 0.01 & 90.47 ± 0.05 \\
		OCTA-500 & UTNet & MICCAI(2021) & \underline{91.32 ± 0.22} & \textbf{91.37 ± 0.75} & 99.37 ± 0.05 & 8.72 ± 0.58 & \underline{\underline{98.83 ± 0.03}} & \underline{90.69 ± 0.23} \\
		3mm & SwinUNet & ECCV(2022) & 88.37 ± 0.04 & 88.02 ± 0.28 & 99.19 ± 0.02 & 11.27 ± 0.26 & 98.44 ± 0.01 &  87.53 ± 0.05 \\
		& OCT2Former & CMPB(2023) & 90.51 ± 0.24 & 89.14 ± 0.92 & \underline{99.43 ± 0.05} & \underline{8.06 ± 0.60} & 98.74 ± 0.02 &  89.84 ± 0.25 \\
		& FRNet & ICASSP(2024) & 90.80 ± 0.09 & 90.61 ± 0.55 & 99.35 ± 0.05 & 9.01 ± 0.57 & 98.76 ± 0.02 & 90.13 ± 0.10 \\
		& DGNet & BSPC(2025) & \underline{\underline{91.26 ± 0.05}} & 90.86 ± 0.49 & 99.40 ± 0.04 & 8.33 ± 0.45 & \underline{98.83 ± 0.01} & \underline{\underline{90.63 ± 0.06}} \\
		\midrule
		& LSENet(Ours) & & \textbf{92.03 ± 0.11} & \underline{91.25 ± 0.61} & \textbf{99.49 ± 0.05} & \textbf{7.17 ± 0.62} & \textbf{98.93 ± 0.02} & \textbf{91.45 ± 0.12} \\
		\hline\midrule
		& UNet & MICCAI(2015) & 88.13 ± 0.04 & 87.01 ± 0.35 & \textbf{98.91 ± 0.04} & \underline{10.73 ± 0.35} & \underline{\underline{97.78 ± 0.01}} & 86.90 ± 0.04 \\
		& CENet & TMI(2019) & 87.38 ± 0.04 & 88.28 ± 0.34 & 98.56 ± 0.05 & 13.51 ± 0.34 & 97.58 ± 0.01 & 86.04 ± 0.04 \\
		& VesselNet & MICCAI(2019) & 87.30 ± 0.10 & 86.01 ± 0.58 & \underline{\underline{98.85 ± 0.07}} & \underline{\underline{11.36 ± 0.56}} & 97.63 ± 0.02 &  86.00 ± 0.11 \\
		& CSNet & MedIA(2021) & \underline{\underline{88.20 ± 0.09}} & \underline{\underline{88.76 ± 0.40}} & 98.69 ± 0.07 & 12.35 ± 0.50 & 97.75 ± 0.03 & 86.96 ± 0.10 \\	
		OCTA-500 & UTNet & MICCAI(2021) & \underline{88.65 ± 0.19} & 88.70 ± 1.22 & 98.81 ± 0.17 & 11.38 ± 1.29 & \underline{97.85 ± 0.05} & \underline{87.46 ± 0.21} \\
		6mm& SwinUNet & ECCV(2022) & 86.31 ± 0.04 & 85.80 ± 0.53 & 98.64 ± 0.07 & 13.17 ± 0.55 & 97.42 ± 0.02 &  84.89 ± 0.04 \\
		& OCT2Former & CMPB(2023) & 87.40 ± 0.27 & 86.96 ± 0.54 & 98.74 ± 0.12 & 12.14 ± 0.93 & 97.63 ± 0.07 &  86.09 ± 0.30 \\
		& FRNet & ICASSP(2024) & 87.59 ± 0.14 & \underline{88.84 ± 0.42} & 98.53 ± 0.07 & 13.63 ± 0.55 & 97.62 ± 0.04 & 86.27 ± 0.16 \\
		& DGNet & BSPC(2025) & 88.17 ± 0.07 & 88.58 ± 0.77 & 98.71 ± 0.11 & 12.23 ± 0.80 & 97.75 ± 0.03 & \underline{\underline{86.92 ± 0.08}} \\
		\midrule
		& LSENet(Ours) & & \textbf{89.14 ± 0.06} & \textbf{88.88 ± 0.52} & \underline{98.90 ± 0.08} & \textbf{10.59 ± 0.59} & \textbf{97.95 ± 0.02} & \textbf{88.01 ± 0.07} \\
		\hline\midrule
		& UNet & MICCAI(2015) & 86.75 ± 0.18 & 89.00 ± 0.62 & \underline{97.41 ± 0.20} & \underline{15.39 ± 0.89} & 96.25 ± 0.08 & 84.56 ± 0.23 \\				
		& CENet & TMI(2019) & 82.27 ± 0.12 & 84.41 ± 0.21 & 96.68 ± 0.09 & 19.76 ± 0.37 & 94.98 ± 0.05 & 79.35 ± 0.15 \\	
		& VesselNet & MICCAI(2019) & 84.52 ± 0.13 & 85.53 ± 0.51 & 97.30 ± 0.10 & 16.47 ± 0.43 & 95.68 ± 0.04 &  82.01 ± 0.15 \\	
		& CSNet & MedIA(2021) & 86.63 ± 0.26 & 90.21 ± 0.33 & 97.11 ± 0.14 & 16.68 ± 0.62 & 96.16 ± 0.09 & 84.39 ± 0.31 \\	
		ROSE-1 & UTNet & MICCAI(2021) & \underline{86.97 ± 0.19} & 90.03 ± 0.84 & 97.28 ± 0.19 & 15.88 ± 0.83 & \underline{96.28 ± 0.07} & \underline{84.80 ± 0.23} \\
		& SwinUNet & ECCV(2022) & 78.28 ± 0.06 & 80.35 ± 0.26 & 96.01 ± 0.06 & 23.69 ± 0.24 & 93.85 ± 0.03 &  74.70 ± 0.07 \\		
		& OCT2Former & CMPB(2023) & \underline{87.78 ± 0.12} & \underline{90.31 ± 0.62} & \underline{97.53 ± 0.17} & \underline{14.61 ± 0.75} & \underline{96.53 ± 0.06} & \underline{85.76 ± 0.16} \\
		& FRNet & ICASSP(2024) & 85.86 ± 0.35 & \underline{\underline{90.29 ± 0.83}} & 96.79 ± 0.31 & 18.14 ± 1.31 & 95.90 ± 0.16 & 83.47 ± 0.45 \\
		& DGNet  & BSPC(2025) & 86.67 ± 0.17 & \textbf{91.42 ± 0.46} & 96.88 ± 0.16 & 17.60 ± 0.67 & 96.12 ± 0.08 & 84.41 ± 0.22 \\
		\midrule
		& LSENet(Ours) & & \textbf{88.22 ± 0.07} & 89.85 ± 0.80 & \textbf{97.79 ± 0.17} & \textbf{13.34 ± 0.82} & \textbf{96.69 ± 0.04} & \textbf{86.30 ± 0.09} \\	
		\hline\midrule
		& UNet & MICCAI(2015) & \underline{\underline{90.60 ± 0.20}} & 89.34 ± 0.67 & 99.09 ± 0.05 & \underline{\underline{8.09 ± 0.34}} & \underline{\underline{98.08 ± 0.03}} & \underline{\underline{89.54 ± 0.22}} \\			
		& CENet & TMI(2019) & 89.10 ± 0.04 & 89.14 ± 0.37 & 98.74 ± 0.06 & 10.94 ± 0.44 & 97.75 ± 0.02 & 87.84 ± 0.05 \\	
		& VesselNet & MICCAI(2019) & 89.34 ± 0.14 & 87.49 ± 0.40 & 99.04 ± 0.05 & 8.72 ± 0.40 & 97.84 ± 0.03 &  88.14 ± 0.15 \\
		& CSNet & MedIA(2021) & 90.04 ± 0.22 & \underline{\underline{90.11 ± 0.56}} & 98.84 ± 0.10 & 10.04 ± 0.73 & 97.94 ± 0.05 & 88.89 ± 0.25 \\
		ROSSA & UTNet & MICCAI(2021) & \underline{91.33 ± 0.17} & \underline{90.60 ± 0.67} & \underline{99.10 ± 0.09} & \underline{7.93 ± 0.72} & \underline{98.22 ± 0.04} & \underline{90.33 ± 0.19} \\
		& SwinUNet & ECCV(2022) & 87.84 ± 0.05 & 87.44 ± 0.38 & 98.66 ± 0.06 & 11.76 ± 0.43 & 97.50 ± 0.02 &  86.44 ± 0.06 \\		
		& OCT2Former & CMPB(2023) & 89.54 ± 0.15 & 87.39 ± 0.97 & \underline{\underline{99.10 ± 0.10}} & 8.19 ± 0.76 & 97.89 ± 0.01 & 88.37 ± 0.15 \\
		& FRNet & ICASSP(2024) & 89.80 ± 0.15 & 89.35 ± 0.49 & 98.89 ± 0.09 & 9.73 ± 0.65 & 97.90 ± 0.04 & 88.63 ± 0.17 \\
		& DGNet & BSPC(2025) & 90.27 ± 0.19 & 89.43 ± 0.78 & 99.00 ± 0.09 & 8.87 ± 0.68 & 98.01 ± 0.04 & 89.16 ± 0.21 \\
		\midrule
		& LSENet(Ours) & & \textbf{91.59 ± 0.15} & \textbf{90.89 ± 0.51} & \textbf{99.13 ± 0.08} & \textbf{7.69 ± 0.62} & \textbf{98.27 ± 0.04} & \textbf{90.63 ± 0.17} \\		
		\bottomrule
	\end{tabular}
	\vspace{-0.55cm}
\end{table*}
\section{Experiments}
\subsection{Datasets and Metrics}

 We evaluate our model on three public OCTA vessel segmentation datasets: ROSE-1 \cite{42}, OCTA-500 \cite{41}, and ROSSA \cite{12}. \textbf{ROSE-1} \cite{42}: Contains 39 images ($304 \times 304$, $3 \times 3 \text{ mm}^2$) of the Superficial Vascular Complex (SVC). We follow the official split (27 Train / 3 Val / 9 Test). \textbf{OCTA-500} \cite{41}: Includes two subsets: OCTA-6M ($6 \times 6 \text{ mm}^2$) and OCTA-3M ($3 \times 3 \text{ mm}^2$). We use the large-vessel annotations from the ILM-OPL maximum projection and follow the official splits: 240/10/50 (Train/Val/Test) for OCTA-6M and 140/10/50 for OCTA-3M. \textbf{ROSSA} \cite{12}: We use the 300 manually annotated images, splitting them into 240/10/50 (Train/Val/Test) following the previous ratio.
 
We utilize six standard metrics to comprehensively evaluate the model's performance from different perspectives:  \textbf{Dice}: The primary metric for evaluating spatial overlap between the prediction and ground truth, especially suitable for imbalanced tasks like vessel segmentation. \textbf{Sensitivity (Sen)}: Measures the ability to correctly identify actual vessels. High Sen indicates a higher target vessel recognition rate. \textbf{Specificity (Spe)}: Measures the ability to correctly identify the background. High Spe indicates good noise suppression and few false alarms. \textbf{Accuracy (Acc)}: The overall correct classification rate for all pixels. \textbf{False Discovery Rate (FDR)}: Measures the proportion of incorrect predictions among all pixels predicted as vessels. Low FDR represents high precision. \textbf{Kappa Coefficient (Kappa)}: A robust metric for agreement that corrects for the possibility of chance agreement. The evaluation dimensions of these six metrics are distinct and complementary.

\subsection{Implementation Details}
All models were trained using an Intel(R) Xeon(R) Gold 6226R CPU and an NVIDIA RTX 3090 GPU for 500 epochs (batch size=2) using the Adam optimizer (LR=$5\text{e}{-4}$, weight decay=$1\text{e}{-4}$) with a polynomial LR decay (power=0.9) and early stopping (patience=100). To prevent overfitting, we applied aggressive data augmentation (random horizontal/vertical flips, $\pm 30^{\circ}$ rotations). The total layer number was set to 4, with a patch size of 20 for the OCTA-500 6mm dataset and 15 for the remaining three. 

\begin{figure}[t]
	\setlength{\abovecaptionskip}{0.cm}
	\setlength{\belowcaptionskip}{-0.5cm}
	\centering
	\includegraphics[width=1\linewidth]{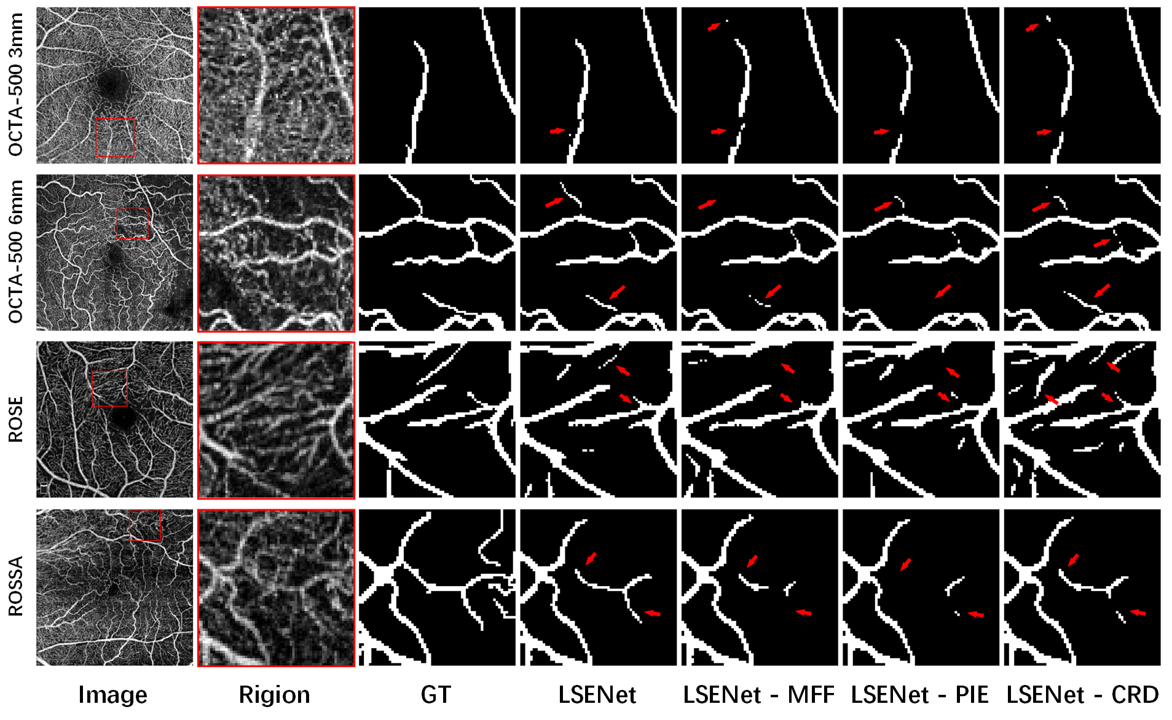}
	\caption{Qualitative results of the ablation study for each module. From top to bottom, each row corresponds to one sample from OCTA-500 3mm, OCTA-500 6mm, ROSE-1, and ROSSA, respectively. Removing MFF, PIE, or CRD leads to distinct failure modes (loss of detail, fragmentation, and poor connectivity).}
	\label{fig:Figure 10}
\end{figure}

\subsection{Comparative Analysis}
We compare LSENet against nine SOTA methods: U-Net \cite{18}, CS-Net \cite{19}, CE-Net \cite{16}, SwinUNet \cite{29}, DGNet \cite{14}, FRNet \cite{12}, Vessel-Net \cite{25}, UTNet \cite{31}, and OCT²former \cite{11}. All models were trained in the same training environment, using the learning rates, loss functions, and optimizers from their respective papers to ensure a fair comparison. We present the mean and standard deviation of the metrics after five consecutive training runs.

Tab.~\ref{tab:Table 1} shows the quantitative results. \textbf{Overall}, our proposed LSENet achieves SOTA performance, ranking first on the vast majority of metrics across all three datasets. The high Dice, FDR, and Kappa metrics indicate that our model can identify target vessels more accurately and reliably, reducing the occurrence of false positives. While it does not top every metric, the exceptions are typically in Sensitivity (Sen) or Specificity (Spe). This is expected, as these two metrics represent a trade-off: a model aggressively optimized to capture every faint vessel (high Sen) may inadvertently increase false positives (lower Spe), and vice versa.

\noindent\textbf{OCTA-500 3mm:} This dataset serves as a standard baseline, featuring a moderate sample size and a balanced segmentation difficulty. On the 3mm dataset, our method performed the best on all metrics except Sen. The Dice, FDR, and Kappa metrics reached \textit{92.03\%}, \textit{7.17\%}, and \textit{91.45\%}, respectively, significantly surpassing the second-place method by \textit{0.71\%}, \textit{0.89\%}, and \textit{0.76\%}. 

\noindent\textbf{OCTA-500 6mm:} While larger in sample size, this dataset presents a greater challenge due to its wider imaging area, which includes a higher density of segmentation targets and more complex structures. On the 6mm dataset, our method demonstrated top-tier performance by ranking first on all metrics except for Spe. This lead was particularly significant in Dice and Kappa, where our model outperformed the second-place method by \textit{0.49\%} and \textit{0.55\%}, respectively.

\noindent\textbf{ROSE-1:} This dataset is significantly smaller, which rigorously tests the model's learning efficiency and its ability to generalize from limited data. On this dataset, our model also achieved the best performance. Apart from the Sen metric, it surpassed the second-place OCT2Former by substantial margins in key metrics, leading in FDR by 1.27\%, Kappa by 0.54\%, and Dice by 0.44\%.

\noindent\textbf{ROSSA:} As a recent dataset, ROSSA is characterized by a higher prevalence of noise, directly challenging the model's noise suppression capabilities and overall robustness. Through the collaboration of the three modules, our model demonstrated superior performance by achieving the best results across all metrics, securing the top rank and establishing a state-of-the-art over the second-place UTNet.

\begin{table}[t]
	\centering
	\scriptsize
	\setlength{\tabcolsep}{2pt}
	
	\caption{Ablation study of the Multiscale Feature Fusion module (MFF), Patch Information Enhance module (PIE), and Connectivity Refinement Decoder (CRD) on the OCTA-500 6mm dataset. (Mean ± SD). The formatting and symbol conventions are the same as those defined in Tab.~\ref{tab:Table 1}. }
	\label{tab:Table 2}
	\begin{tabular}{ccccccccc}
		\toprule
		MFF & PIE & CRD & Dice\% ↑ & Sen\% ↑ & Spe\% ↑ & FDR\% ↓ & Acc\% ↑ & Kappa\% ↑ \\
		\midrule
		× & × & × & 88.49 & 88.72 & 98.77 & 11.73 & 97.81 & 87.28 \\		
		× & \checkmark & × & 88.81 & 88.73 & \underline{\underline{98.84}} & \underline{\underline{11.10}} & 97.88 & 87.64 \\	
		\checkmark & × & × & 88.89 & \underline{\underline{89.22}} & 98.79 & 11.43 & 97.89 & 87.72 \\
		× & × & \checkmark & 88.85 & 89.15 & 98.80 & 11.43 & 97.88 & 87.68 \\		
		\checkmark & × & \checkmark & \underline{89.01} & 88.62 & \textbf{98.90} & \underline{10.60} & \underline{97.93} &  \underline{87.86} \\
		× & \checkmark & \checkmark & 88.96 & \underline{89.38} & 98.79 & 11.46 & \underline{\underline{97.90}} &  87.80 \\	
		\checkmark & \checkmark & × & \underline{\underline{88.98}} & \textbf{89.42} & 98.79 & 11.44 & 97.90 & \underline{\underline{87.82}} \\
		\checkmark & \checkmark & \checkmark & \textbf{89.14} & 88.88 & \underline{98.90} & \textbf{10.59} & \textbf{97.95} & \textbf{88.01} \\	
		\bottomrule
	\end{tabular}
	\vspace{-0.55cm}
\end{table}

\subsection{Ablation Study}
\label{subsec:Ablation Study}

We conducted a thorough ablation study to validate the contribution of each proposed module. Tab.~\ref{tab:Table 2} presents the results of experiments performed on the OCTA-500 6mm dataset, selected for its larger data volume and greater segmentation difficulty. Starting from a baseline model (a U-Net with channels fixed at 64), the metrics quantitatively demonstrate the performance improvement provided by each individual module. Fig.~\ref{fig:Figure 10} provides a qualitative visualization of the specific contribution of each proposed module. For a clear comparison, we illustrate the segmentation results of the full LSENet against variants where each module has been individually removed.

PIE (Patch Information Enhance module): This module focuses on regional feature selection. It selectively enhances features in challenging areas, such as low-contrast regions, while suppressing background noise. When PIE is removed, the model's perceptual ability in these low-contrast areas drops sharply, manifesting as large segments of the target vasculature breaking or disappearing(Fig.~\ref{fig:Figure 10} the second-to-last column).

MFF (Multiscale Feature Fusion module): This module enriches the feature representation. Without MFF, the model's ability to perceive vessels at different scales noticeably degrades, leading to weakened identification of finer vessels (Fig.~\ref{fig:Figure 10} the third-to-last column). While the inclusion of MFF might introduce more potential candidates by increasing sensitivity, it fundamentally enhances the model's foundational perceptual capability.

CRD (Connectivity Refinement Decoder): This module acts as a refinement stage. Leveraging its large receptive field ($11 \times 11$ kernel), it makes a more ``cautious'' and context-aware decision when producing the final output. Without CRD, the segmentation results become fragmented, connectivity decreases, and false positives increase(Fig.~\ref{fig:Figure 10} the last column).

In summary, the MFF and PIE modules work synergistically to significantly boost the model's perception and recall of diverse and low-contrast targets. The CRD then screens and refines these proposals, effectively filtering artifacts and enforcing vascular connectivity. Each module fulfills a distinct role, enabling LSENet to achieve high performance in complex scenarios.

\begin{figure}[t]
	\setlength{\abovecaptionskip}{0.cm}
	\setlength{\belowcaptionskip}{-0.5cm}
	\centering
	\includegraphics[width=1\linewidth]{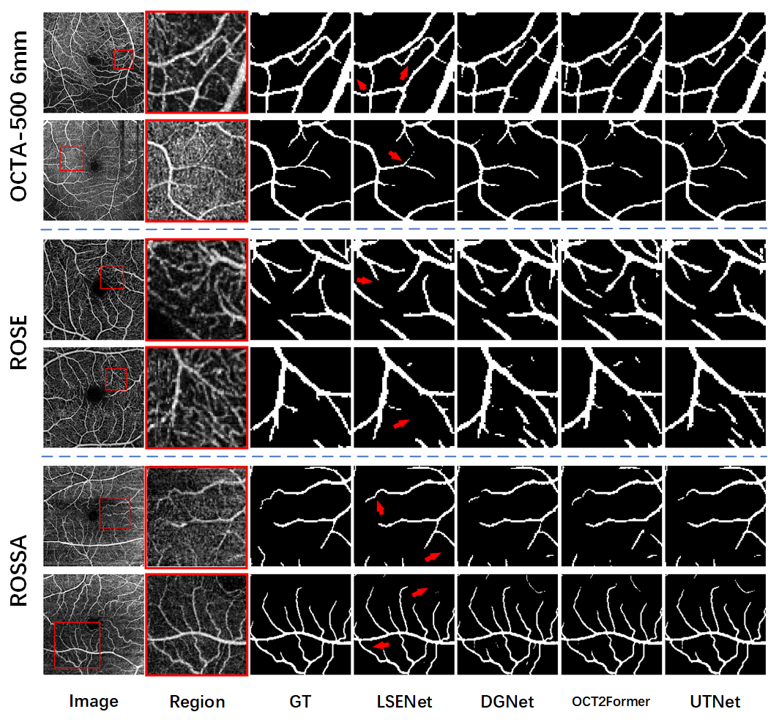}
	\caption{Qualitative comparison on samples from OCTA-500 6mm, ROSE-1, and ROSSA. Columns (from left to right) show: Original Image, Magnified Region, Ground Truth (GT), LSENet (Ours), and competing methods. Red arrows highlight challenging regions where LSENet excels at maintaining vessel connectivity and suppressing false positives. }
	\label{fig:Figure 9}
\end{figure}

\subsection{Discussion}
\label{subsec:Discussion}
\textbf{What roles do the MFF, PIE, and CRD modules play?} The role of PIE is to use local attention to enhance local features. However, we found that the performance of the PIE module was constrained by the encoder's capabilities. Therefore, we added the MFF module to enable the model to acquire richer information, but this also led to an excess of fragmented targets in our final output. Consequently, we modified the decoder into the CRD, which allows the model to treat the features captured by the preceding modules more cautiously, enhancing the connectivity of the final result.

\noindent	
\textbf{Why was a more aggressive data augmentation strategy adopted?} We increased the rotation range from $\pm$10° to $\pm$30° to create a more challenging and diverse training distribution, preventing larger models from overfitting early and ensuring a fair comparison of the true learning potential across all architectures. We emphasize that all models, including the traditional CNN-based baselines, were trained from scratch under this identical strategy and monitored until full convergence. By subjecting every model to the same rigorous data regime, we aimed to demonstrate that LSENet’s performance gains stem from its intrinsic architectural capacity to capture robust vessel features rather than from dataset bias. This ensures that the observed SOTA results reflect the models' actual generalization capabilities in complex clinical environments.

\noindent	
\textbf{What are LSENet's advantages compared to other methods?} Fig.~\ref{fig:Figure 9} demonstrates that our model effectively handles low-contrast regions and produces segmentation results with superior topological connectivity. Architecturally, LSENet enhances interpretability by utilizing downsampled raw inputs across layers and a large-receptive-field final convolution to capture critical vascular structures. Furthermore, as shown in Tab.~\ref{tab:Table 3}, our model achieves an exceptionally compact footprint. With only 2.08M parameters, it is approximately 8$\times$ smaller than U-Net (17.3M), 27$\times$ smaller than UTNet (57.5M), and 13$\times$ smaller than SwinUNet (27.1M). This high parameter efficiency demonstrates that LSENet can achieve state-of-the-art representation power while significantly reducing storage requirements, making it highly suitable for memory-constrained medical imaging environments.
\begin{table}[t]
	\setlength{\abovecaptionskip}{0.cm}
	\setlength{\belowcaptionskip}{-0.5cm}
	\centering
	\small
	\caption{Comparison of parameters, FLOPs, and time for retinal vessel segmentation under $224 \times 224$ input conditions.}
	\label{tab:Table 3}
	\begin{tabular}{cccc}
		\toprule
		Model & Params/M & FLOPs/G & Time/ms \\
		\midrule
		UNet & 17.27 & 30.71 & 4.11\\
		UTNet & 57.45 & 62.23 & 12.15\\
		SwinUNet & 27.14 & 5.90 & 11.19\\
		OCT2Former & 7.35 & 50.07 & 26.88\\
		FRNet & 0.01 & 18.63 & 6.92\\
		DGNet & 1.04 & 2.74 & 20.61\\
		\hline
		LSENet(Ours) & 2.08 & 55.49 & 27.19\\
		\bottomrule
	\end{tabular}
	\vspace{-0.55cm}
\end{table}

\section{Conclusion}
\label{sec:conclusion}

This paper presents LSENet, a lightweight architecture designed to tackle low local contrast and vessel discontinuity in OCTA segmentation via MFF, PIE, and CRD modules. Extensive evaluations on three public datasets demonstrate that LSENet achieves state-of-the-art performance with superior topological connectivity and noise robustness. With its parameter efficiency and effective local feature enhancement, LSENet offers a promising solution for automated clinical diagnosis in OCTA analysis.
\section*{Acknowledgments}
This work was supported by the National Science Foundation of China [U23A20389, 62306095, 82441009, 82441008, 62506101]; the Heilongjiang Natural Science Foundation of China [LH2024F021]; the China Postdoctoral Science Foundation [2024M764190]; the Heilongjiang Postdoctoral Science Foundation [LBH-Z24180]; the Heilongjiang Province Science and Technology Talent Support Program Project [CYQN24030]; the Fundamental Research Funds for the Central Universities [HIT.NSFJG202439, HIT.NSFJG202434].

{
	\small
	\bibliographystyle{ieeenat_fullname}
	\bibliography{main}

\begin{thebibliography}{48}
\providecommand{\natexlab}[1]{#1}
\providecommand{\url}[1]{\texttt{#1}}
\expandafter\ifx\csname urlstyle\endcsname\relax
  \providecommand{\doi}[1]{doi: #1}\else
  \providecommand{\doi}{doi: \begingroup \urlstyle{rm}\Url}\fi

\bibitem[Cao et~al.(2025)Cao, Peng, Zhou, Wu, Zhang, and Yan]{38}
Guogang Cao, Zeyu Peng, Zhilin Zhou, Yan Wu, Yunqing Zhang, and Rugang Yan.
\newblock Multi-task octa image segmentation with innovative dimension
  compression.
\newblock \emph{Pattern Recognition}, 159:\penalty0 111123, 2025.

\bibitem[Cao et~al.(2022)Cao, Wang, Chen, Jiang, Zhang, Tian, and Wang]{29}
Hu Cao, Yueyue Wang, Joy Chen, Dongsheng Jiang, Xiaopeng Zhang, Qi Tian, and
  Manning Wang.
\newblock Swin-unet: Unet-like pure transformer for medical image
  segmentation.
\newblock In \emph{Computer Vision – ECCV 2022 Workshops}, pages 205--218.
  Springer Nature Switzerland, 2022.

\bibitem[Carpineto et~al.(2016)Carpineto, Mastropasqua, Marchini, Toto,
  Di~Nicola, and Di~Antonio]{4}
P. Carpineto, R. Mastropasqua, G. Marchini, L. Toto, M. Di~Nicola, and L.
  Di~Antonio.
\newblock Reproducibility and repeatability of foveal avascular zone
  measurements in healthy subjects by optical coherence tomography angiography.
\newblock \emph{Br J Ophthalmol}, 100\penalty0 (5):\penalty0 671--6, 2016.

\bibitem[Chu et~al.(2021)Chu, Tian, Wang, Zhang, Ren, Wei, Xia, and Shen]{44}
Xiangxiang Chu, Zhi Tian, Yuqing Wang, Bo Zhang, Haibing Ren, Xiaolin Wei,
  Huaxia Xia, and Chunhua Shen.
\newblock Twins: revisiting the design of spatial attention in vision
  transformers.
\newblock \emph{Nips '21}, 2021.

\bibitem[Dosovitskiy et~al.(2021)Dosovitskiy, Beyer, Kolesnikov, Weissenborn,
  Zhai, Unterthiner, Dehghani, Minderer, Heigold, Gelly, Uszkoreit, and
  Houlsby]{43}
Alexey Dosovitskiy, Lucas Beyer, Alexander Kolesnikov, Dirk Weissenborn,
  Xiaohua Zhai, Thomas Unterthiner, Mostafa Dehghani, Matthias Minderer, Georg
  Heigold, Sylvain Gelly, Jakob Uszkoreit, and Neil Houlsby.
\newblock An image is worth 16x16 words: Transformers for image recognition at
  scale.
\newblock In \emph{2021 International Conference on Learning Representations
  (ICLR)}, page arXiv 2010.11929, 2021.

\bibitem[Fu et~al.(2016)Fu, Xu, Lin, Kee~Wong, and Liu]{20}
Huazhu Fu, Yanwu Xu, Stephen Lin, Damon~Wing Kee~Wong, and Jiang Liu.
\newblock Deepvessel: Retinal vessel segmentation via deep learning and
  conditional random field.
\newblock In \emph{Medical Image Computing and Computer-Assisted Intervention
  – MICCAI 2016}, pages 132--139. Springer International Publishing, 2016.

\bibitem[Gao et~al.(2021)Gao, Zhou, and Metaxas]{31}
Yunhe Gao, Mu Zhou, and Dimitris~N. Metaxas.
\newblock Utnet: A hybrid transformer architecture for medical image
  segmentation.
\newblock In \emph{Medical Image Computing and Computer Assisted Intervention
  – MICCAI 2021}, pages 61--71. Springer International Publishing, 2021.

\bibitem[Gu et~al.(2019)Gu, Cheng, Fu, Zhou, Hao, Zhao, Zhang, Gao, and
  Liu]{16}
Z. Gu, J. Cheng, H. Fu, K. Zhou, H. Hao, Y. Zhao, T. Zhang, S. Gao, and J. Liu.
\newblock Ce-net: Context encoder network for 2d medical image segmentation.
\newblock \emph{IEEE Transactions on Medical Imaging}, 38\penalty0
  (10):\penalty0 2281--2292, 2019.

\bibitem[Hu et~al.(2022)Hu, Jiang, Zhang, Li, and Gao]{40}
K. Hu, S. Jiang, Y. Zhang, X. Li, and X. Gao.
\newblock Joint-seg: Treat foveal avascular zone and retinal vessel
  segmentation in octa images as a joint task.
\newblock \emph{IEEE Transactions on Instrumentation and Measurement},
  71:\penalty0 1--13, 2022.

\bibitem[Jiang et~al.(2026)Jiang, Li, Li, Xing, Yu, Xie, and Ta]{49}
Tao Jiang, Ying Li, Yifang Li, Wenyu Xing, Ming Yu, Feng Xie, and Dean Ta.
\newblock A segmentation knowledge-based global-local attention network for
  tumor classification in breast ultrasound images.
\newblock \emph{Pattern Recognition}, 171, 2026.

\bibitem[Jin et~al.(2019)Jin, Meng, Pham, Chen, Wei, and Su]{21}
Qiangguo Jin, Zhaopeng Meng, Tuan~D. Pham, Qi Chen, Leyi Wei, and Ran Su.
\newblock Dunet: A deformable network for retinal vessel segmentation.
\newblock \emph{Knowledge-Based Systems}, 178:\penalty0 149--162, 2019.

\bibitem[Kashani et~al.(2017)Kashani, Chen, Gahm, Zheng, Richter, Rosenfeld,
  Shi, and Wang]{2}
Amir~H. Kashani, Chieh-Li Chen, Jin~K. Gahm, Fang Zheng, Grace~M. Richter,
  Philip~J. Rosenfeld, Yonggang Shi, and Ruikang~K. Wang.
\newblock Optical coherence tomography angiography: A comprehensive review of
  current methods and clinical applications.
\newblock \emph{Progress in Retinal and Eye Research}, 60:\penalty0 66--100,
  2017.

\bibitem[Kreitner et~al.(2024)Kreitner, Paetzold, Rauch, Chen, Hagag, Fayed,
  Sivaprasad, Rausch, Weichsel, Menze, Harders, Knier, Rueckert, and
  Menten]{36}
L. Kreitner, J.~C. Paetzold, N. Rauch, C. Chen, A.~M. Hagag, A.~E. Fayed, S.
  Sivaprasad, S. Rausch, J. Weichsel, B.~H. Menze, M. Harders, B. Knier, D.
  Rueckert, and M.~J. Menten.
\newblock Synthetic optical coherence tomography angiographs for detailed
  retinal vessel segmentation without human annotations.
\newblock \emph{IEEE Transactions on Medical Imaging}, 43\penalty0
  (6):\penalty0 2061--2073, 2024.

\bibitem[L~Srinidhi et~al.(2017)L~Srinidhi, Aparna, and Rajan]{9}
Chetan L~Srinidhi, P. Aparna, and Jeny Rajan.
\newblock Recent advancements in retinal vessel segmentation.
\newblock \emph{Journal of Medical Systems}, 41\penalty0 (4):\penalty0 70,
  2017.

\bibitem[Laíns et~al.(2021)Laíns, Wang, Cui, Katz, Vingopoulos, Staurenghi,
  Vavvas, Miller, and Miller]{8}
Inês Laíns, Jay~C. Wang, Ying Cui, Raviv Katz, Filippos Vingopoulos, Giovanni
  Staurenghi, Demetrios~G. Vavvas, Joan~W. Miller, and John~B. Miller.
\newblock Retinal applications of swept source optical coherence tomography
  (oct) and optical coherence tomography angiography (octa).
\newblock \emph{Progress in Retinal and Eye Research}, 84:\penalty0 100951,
  2021.

\bibitem[Li et~al.(2020)Li, Chen, Ji, Xie, Yuan, Chen, and Li]{37}
M. Li, Y. Chen, Z. Ji, K. Xie, S. Yuan, Q. Chen, and S. Li.
\newblock Image projection network: 3d to 2d image segmentation in octa images.
\newblock \emph{IEEE Transactions on Medical Imaging}, 39\penalty0
  (11):\penalty0 3343--3354, 2020.

\bibitem[Li et~al.(2024)Li, Huang, Xu, Yang, Zhang, Ji, Xie, Yuan, Liu, and
  Chen]{41}
Mingchao Li, Kun Huang, Qiuzhuo Xu, Jiadong Yang, Yuhan Zhang, Zexuan Ji, Keren
  Xie, Songtao Yuan, Qinghuai Liu, and Qiang Chen.
\newblock Octa-500: A retinal dataset for optical coherence tomography
  angiography study.
\newblock \emph{Medical Image Analysis}, 93:\penalty0 103092, 2024.

\bibitem[Li et~al.(2025)Li, Zhang, Zhao, Shi, and Zhou]{14}
Zhenli Li, Xinpeng Zhang, Meng Zhao, Fan Shi, and Wei Zhou.
\newblock Direction-guided network for retinal vessel segmentation in octa
  images.
\newblock \emph{Biomedical Signal Processing and Control}, 103:\penalty0
  107455, 2025.

\bibitem[Liu et~al.(2023{\natexlab{a}})Liu, Hirota, and Dai]{47}
Chang Liu, Kaoru Hirota, and Yaping Dai.
\newblock Patch attention convolutional vision transformer for facial
  expression recognition with occlusion.
\newblock \emph{Information Sciences}, 619:\penalty0 781--794,
  2023{\natexlab{a}}.

\bibitem[Liu et~al.(2024)Liu, Zhao, Shen, Geng, Zhang, Yang, and Zhang]{22}
Jianhua Liu, Dongxin Zhao, Juncai Shen, Peng Geng, Ying Zhang, Jiaxin Yang, and
  Ziqian Zhang.
\newblock Hrd-net: High resolution segmentation network with adaptive learning
  ability of retinal vessel features.
\newblock \emph{Computers in Biology and Medicine}, 173:\penalty0 108295, 2024.

\bibitem[Liu et~al.(2023{\natexlab{b}})Liu, Zhang, Yao, and Tang]{10}
Xiaoming Liu, Di Zhang, Junping Yao, and Jinshan Tang.
\newblock Transformer and convolutional based dual branch network for retinal
  vessel segmentation in octa images.
\newblock \emph{Biomedical Signal Processing and Control}, 83:\penalty0 104604,
  2023{\natexlab{b}}.

\bibitem[Liu et~al.(2025)Liu, Shen, Zhong, Xiong, and Chen]{15}
Xinyi Liu, Hailan Shen, Wenyan Zhong, Wanqing Xiong, and Zailiang Chen.
\newblock Dsdc-net: Semi-supervised superficial octa vessel segmentation for
  false positive reduction.
\newblock \emph{Pattern Recognition}, 165:\penalty0 111592, 2025.

\bibitem[Liu et~al.(2021)Liu, Lin, Cao, Hu, Wei, Zhang, Lin, and Guo]{30}
Z. Liu, Y. Lin, Y. Cao, H. Hu, Y. Wei, Z. Zhang, S. Lin, and B. Guo.
\newblock Swin transformer: Hierarchical vision transformer using shifted
  windows.
\newblock In \emph{2021 IEEE/CVF International Conference on Computer Vision
  (ICCV)}, pages 9992--10002, 2021.

\bibitem[Ma et~al.(2021)Ma, Hao, Xie, Fu, Zhang, Yang, Wang, Liu, Zheng, and
  Zhao]{42}
Y. Ma, H. Hao, J. Xie, H. Fu, J. Zhang, J. Yang, Z. Wang, J. Liu, Y. Zheng, and
  Y. Zhao.
\newblock Rose: A retinal oct-angiography vessel segmentation dataset and new
  model.
\newblock \emph{IEEE Transactions on Medical Imaging}, 40\penalty0
  (3):\penalty0 928--939, 2021.

\bibitem[Moons and De~Groef(2022)]{6}
Lieve Moons and Lies De~Groef.
\newblock Multimodal retinal imaging to detect and understand alzheimer’s and
  parkinson’s disease.
\newblock \emph{Current Opinion in Neurobiology}, 72:\penalty0 1--7, 2022.

\bibitem[Mou et~al.(2021)Mou, Zhao, Fu, Liu, Cheng, Zheng, Su, Yang, Chen,
  Frangi, Akiba, and Liu]{19}
Lei Mou, Yitian Zhao, Huazhu Fu, Yonghuai Liu, Jun Cheng, Yalin Zheng, Pan Su,
  Jianlong Yang, Li Chen, Alejandro~F. Frangi, Masahiro Akiba, and Jiang Liu.
\newblock Cs2-net: Deep learning segmentation of curvilinear structures in
  medical imaging.
\newblock \emph{Medical Image Analysis}, 67:\penalty0 101874, 2021.

\bibitem[Ning et~al.(2024)Ning, Wang, Chen, and Li]{12}
H. Ning, C. Wang, X. Chen, and S. Li.
\newblock An accurate and efficient neural network for octa vessel segmentation
  and a new dataset.
\newblock In \emph{ICASSP 2024 - 2024 IEEE International Conference on
  Acoustics, Speech and Signal Processing (ICASSP)}, pages 1966--1970, 2024.

\bibitem[Peng et~al.(2025)Peng, He, Wei, Wen, and Hu]{45}
R. Peng, H. He, Y. Wei, Y. Wen, and D. Hu.
\newblock Patch matters: Training-free fine-grained image caption enhancement
  via local perception.
\newblock In \emph{2025 IEEE/CVF Conference on Computer Vision and Pattern
  Recognition (CVPR)}, pages 3963--3973, 2025.

\bibitem[Pujari et~al.(2021)Pujari, Bhaskaran, Sharma, Singh, Phuljhele,
  Saxena, and Azad]{5}
Amar Pujari, Karthika Bhaskaran, Pradeep Sharma, Pallavi Singh, Swati
  Phuljhele, Rohit Saxena, and Shorya~Vardhan Azad.
\newblock Optical coherence tomography angiography in neuro-ophthalmology:
  Current clinical role and future perspectives.
\newblock \emph{Survey of Ophthalmology}, 66\penalty0 (3):\penalty0 471--481,
  2021.

\bibitem[Quan et~al.(2025)Quan, Hou, Yin, and Zhang]{39}
Xiongwen Quan, Guangyao Hou, Wenya Yin, and Han Zhang.
\newblock A multi-modal and multi-stage fusion enhancement network for
  segmentation based on oct and octa images.
\newblock \emph{Information Fusion}, 113:\penalty0 102594, 2025.

\bibitem[Ren et~al.(2024)Ren, Zhou, LIN, Huang, FU, Song, and
  Cheng]{ren2024spikepoint}
Hongwei Ren, Yue Zhou, Xiaopeng LIN, Yulong Huang, Haotian FU, Jie Song, and
  Bojun Cheng.
\newblock Spikepoint: An efficient point-based spiking neural network for event
  cameras action recognition.
\newblock In \emph{The Twelfth International Conference on Learning
  Representations}, 2024.

\bibitem[Ren et~al.(2025)Ren, Zhou, Zhu, Lin, Fu, Huang, Fang, Ma, Yu, and
  Cheng]{10946204}
Hongwei Ren, Yue Zhou, Jiadong Zhu, Xiaopeng Lin, Haotian Fu, Yulong Huang,
  Yuetong Fang, Fei Ma, Hao Yu, and Bojun Cheng.
\newblock Rethinking efficient and effective point-based networks for event
  camera classification and regression, 2025.

\bibitem[Ronneberger et~al.(2015)Ronneberger, Fischer, and Brox]{18}
Olaf Ronneberger, Philipp Fischer, and Thomas Brox.
\newblock U-net: Convolutional networks for biomedical image segmentation.
\newblock In \emph{Medical Image Computing and Computer-Assisted Intervention
  – MICCAI 2015}, pages 234--241. Springer International Publishing, 2015.

\bibitem[Sampson et~al.(2022)Sampson, Dubis, Chen, Zawadzki, and Sampson]{3}
Danuta~M. Sampson, Adam~M. Dubis, Fred~K. Chen, Robert~J. Zawadzki, and
  David~D. Sampson.
\newblock Towards standardizing retinal optical coherence tomography
  angiography: a review.
\newblock \emph{Light: Science \and Applications}, 11\penalty0 (1):\penalty0
  63, 2022.

\bibitem[Shen et~al.(2024)Shen, Tang, Li, Duan, and Chen]{35}
Hailan Shen, Zheng Tang, Yajing Li, Xuanchu Duan, and Zailiang Chen.
\newblock Haic-net: Semi-supervised octa vessel segmentation with
  self-supervised pretext task and dual consistency training.
\newblock \emph{Pattern Recognition}, 151:\penalty0 110429, 2024.

\bibitem[Shin et~al.(2019)Shin, Nam, Lee, Lim, Lee, Jo, and Kim]{7}
Y.~I. Shin, K.~Y. Nam, S.~E. Lee, H.~B. Lim, M.~W. Lee, Y.~J. Jo, and J.~Y.
  Kim.
\newblock Changes in peripapillary microvasculature and retinal thickness in
  the fellow eyes of patients with unilateral retinal vein occlusion: An octa
  study.
\newblock \emph{Invest Ophthalmol Vis Sci}, 60\penalty0 (2):\penalty0 823--829,
  2019.

\bibitem[Suzuki et~al.(2016)Suzuki, Hirano, Yoshida, Tomiyasu, Uemura,
  Yasukawa, and Ogura]{1}
Norihiro Suzuki, Yoshio Hirano, Munenori Yoshida, Taneto Tomiyasu, Akiyoshi
  Uemura, Tsutomu Yasukawa, and Yuichiro Ogura.
\newblock Microvascular abnormalities on optical coherence tomography
  angiography in macular edema associated with branch retinal vein occlusion.
\newblock \emph{American Journal of Ophthalmology}, 161:\penalty0 126--132.e1,
  2016.

\bibitem[Tan et~al.(2023)Tan, Chen, Meng, Shi, Xiang, Chen, Pan, and Zhu]{11}
Xiao Tan, Xinjian Chen, Qingquan Meng, Fei Shi, Dehui Xiang, Zhongyue Chen,
  Lingjiao Pan, and Weifang Zhu.
\newblock Oct2former: A retinal oct-angiography vessel segmentation
  transformer.
\newblock \emph{Computer Methods and Programs in Biomedicine}, 233:\penalty0
  107454, 2023.

\bibitem[Vaswani et~al.(2017)Vaswani, Shazeer, Parmar, Uszkoreit, Jones, Gomez,
  Kaiser, and Polosukhin]{28}
Ashish Vaswani, Noam Shazeer, Niki Parmar, Jakob Uszkoreit, Llion Jones,
  Aidan~N. Gomez, \L{}ukasz Kaiser, and Illia Polosukhin.
\newblock Attention is all you need.
\newblock In \emph{Proceedings of the 31st International Conference on Neural
  Information Processing Systems}, page 6000–6010, Red Hook, NY, USA, 2017.
  Curran Associates Inc.

\bibitem[Wang et~al.(2023)Wang, Ning, Chen, and Li]{32}
C. Wang, H. Ning, X. Chen, and S. Li.
\newblock Db-unet: Mlp based dual branch unet for accurate vessel segmentation
  in octa images.
\newblock In \emph{ICASSP 2023 - 2023 IEEE International Conference on
  Acoustics, Speech and Signal Processing (ICASSP)}, pages 1--5, 2023.

\bibitem[Wang et~al.(2024)Wang, Chen, Ning, and Li]{13}
C. Wang, X. Chen, H. Ning, and S. Li.
\newblock Sam-octa: A fine-tuning strategy for applying foundation model octa
  image segmentation tasks.
\newblock In \emph{ICASSP 2024 - 2024 IEEE International Conference on
  Acoustics, Speech and Signal Processing (ICASSP)}, pages 1771--1775, 2024.

\bibitem[Wu et~al.(2021)Wu, Wang, Zhong, Lei, Wen, and Qin]{27}
Huisi Wu, Wei Wang, Jiafu Zhong, Baiying Lei, Zhenkun Wen, and Jing Qin.
\newblock Scs-net: A scale and context sensitive network for retinal vessel
  segmentation.
\newblock \emph{Medical Image Analysis}, 70:\penalty0 102025, 2021.

\bibitem[Wu et~al.(2019)Wu, Xia, Song, Zhang, Liu, Zhang, and Cai]{25}
Yicheng Wu, Yong Xia, Yang Song, Donghao Zhang, Dongnan Liu, Chaoyi Zhang, and
  Weidong Cai.
\newblock Vessel-net: Retinal vessel segmentation under multi-path supervision.
\newblock In \emph{Medical Image Computing and Computer Assisted Intervention
  – MICCAI 2019}, pages 264--272. Springer International Publishing, 2019.

\bibitem[Wu et~al.(2020)Wu, Xia, Song, Zhang, and Cai]{26}
Yicheng Wu, Yong Xia, Yang Song, Yanning Zhang, and Weidong Cai.
\newblock Nfn+: A novel network followed network for retinal vessel
  segmentation.
\newblock \emph{Neural Networks}, 126:\penalty0 153--162, 2020.

\bibitem[Ye et~al.(2022)Ye, Pan, Wu, Wang, and Xia]{23}
Y. Ye, C. Pan, Y. Wu, S. Wang, and Y. Xia.
\newblock Mfi-net: Multiscale feature interaction network for retinal vessel
  segmentation.
\newblock \emph{IEEE Journal of Biomedical and Health Informatics}, 26\penalty0
  (9):\penalty0 4551--4562, 2022.

\bibitem[Yuan et~al.(2022)Yuan, Zhang, Wang, and Huang]{24}
Y. Yuan, L. Zhang, L. Wang, and H. Huang.
\newblock Multi-level attention network for retinal vessel segmentation.
\newblock \emph{IEEE Journal of Biomedical and Health Informatics}, 26\penalty0
  (1):\penalty0 312--323, 2022.

\bibitem[Zhu et~al.(2021)Zhu, Sun, Huang, Han, and Zhang]{48}
W. Zhu, L. Sun, J. Huang, L. Han, and D. Zhang.
\newblock Dual attention multi-instance deep learning for alzheimer's disease
  diagnosis with structural mri.
\newblock \emph{IEEE Trans Med Imaging}, 40\penalty0 (9):\penalty0 2354--2366,
  2021.

\bibitem[Zuo et~al.(2025)Zuo, Pang, Hu, Kang, Lv, and Lyu]{46}
Yang Zuo, Chen Pang, Chunyu Hu, Chunmeng Kang, Hongbin Lv, and Lei Lyu.
\newblock Global partition with local enhancement network for multitask
  learning of malignant melanoma.
\newblock \emph{Biomedical Signal Processing and Control}, 106, 2025.

\end{thebibliography}
}



\end{document}